\documentclass[letterpaper, 10pt, conference]{ieeeconf}
\usepackage{array}
\usepackage{float}
\usepackage{amsmath}
\usepackage{amssymb}
\usepackage{graphicx}
\usepackage{algorithm}
\usepackage{algpseudocode}
\usepackage{subfigure}
\usepackage{lmodern}
\usepackage{arydshln}
\usepackage{lipsum}
\usepackage[caption=false,font=footnotesize]{subfig}
\makeatletter

\IEEEoverridecommandlockouts                              
\makeatother
\begin{document}

\title{Detecting and Tracking Small Moving Objects in Wide Area Motion Imagery (WAMI) Using Convolutional Neural Networks (CNNs)}

\author{Yifan Zhou$^{\dagger}$ and Simon Maskell$^{\dagger}$\thanks{$\dagger$ Yifan Zhou and Simon Maskell are with the Department of Electrical Engineering and Electronics, University of Liverpool, Liverpool, UK.   Emails: \{yifanz, smaskell\}@liverpool.ac.uk}
\thanks{$^*$ The matlab implementation with trained network can be found at: \newline \scriptsize https://github.com/zhouyifan233/MovingObjDetector-WAMI.matlab}}

\maketitle
\begin{abstract}
This paper proposes an approach to detect moving objects in Wide Area Motion Imagery (WAMI), in which the objects are both small and well separated. Identifying the objects only using foreground appearance is difficult since a $100-$pixel vehicle is hard to distinguish from objects comprising the background. Our approach is based on background subtraction as an efficient and unsupervised method that is able to output the shape of objects. In order to reliably detect low contrast and small objects, we configure the background subtraction to extract foreground regions that might be objects of interest. While this dramatically increases the number of false alarms, a Convolutional Neural Network (CNN) considering both spatial and temporal information is then trained to reject the false alarms. In areas with heavy traffic, the background subtraction yields merged detections. To reduce the complexity of multi-target tracker needed, we train another CNN to predict the positions of multiple moving objects in an area. Our approach shows competitive detection performance on smaller objects relative to the state-of-the-art. We adopt a GM-PHD filter to associate detections over time and analyse the resulting performance.
\end{abstract}

\begin{keywords}
Wide Area Motion Imagery, Moving object detection, Background subtraction, Convolutional Neural Networks, Multi-target tracking
\end{keywords}

\IEEEpeerreviewmaketitle{}

\section{Introduction} \label{sec:introductoin}

\begin{figure*}[t]
\centering
\subfigure[The input image.]{
\includegraphics[width=5cm]{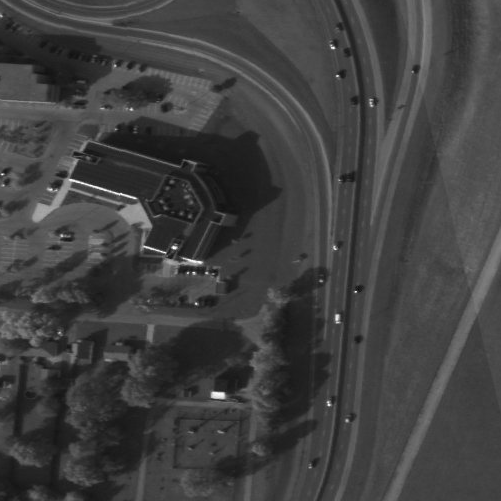}
}
\subfigure[The result of background subtraction after image opening.]{
\includegraphics[width=5cm]{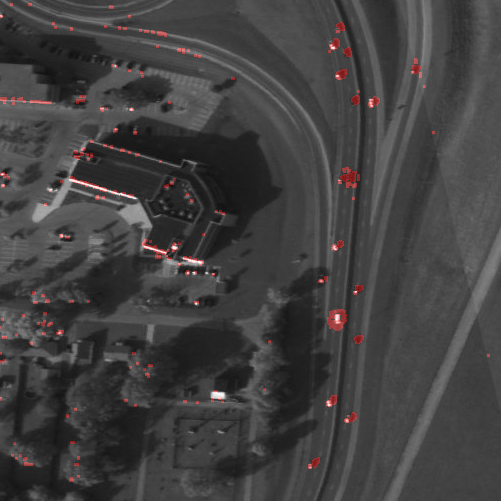}
}
\subfigure[The centres of the windows to be predicted by the classification CNN.]{
\includegraphics[width=5cm]{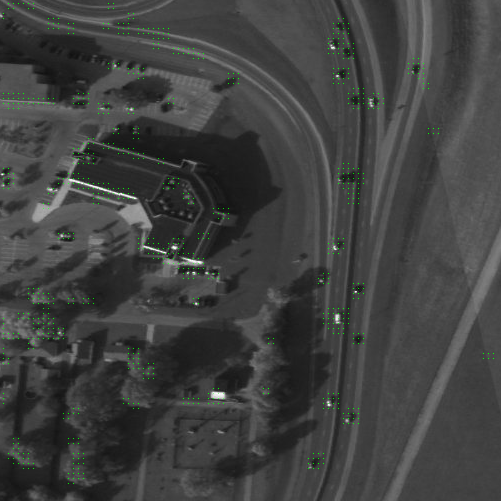}
}
\subfigure[The accepted regions of detections by the classification CNN.]{
\includegraphics[width=5cm]{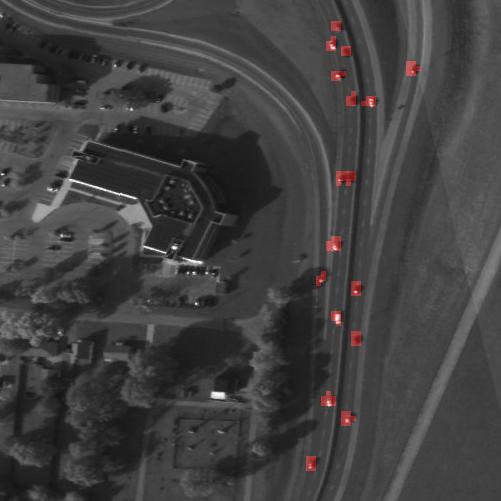}
}
\subfigure[The red detections are accepted directly. The green detections will go through the regression CNN.]{
\includegraphics[width=5cm]{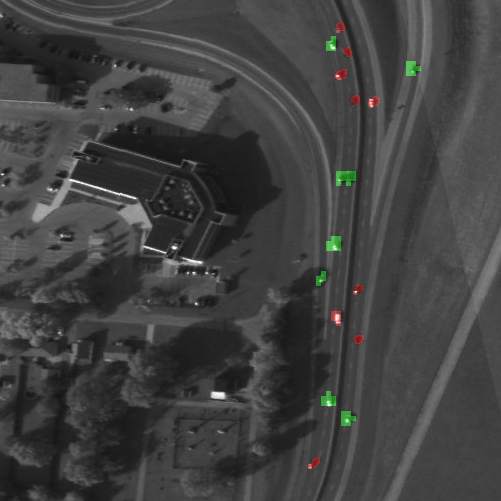}
}
\subfigure[The final detection result. The detections are cropped out by green polygons. The red points are from the ground-truth.]{
\includegraphics[width=5cm]{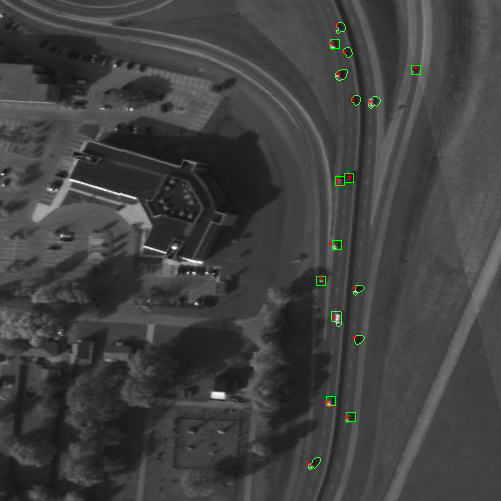}
}
\caption{An example of the processing chain of the proposed moving object detector. This example is a sub-set of AOI 34.\label{fig:example-detector}}
\end{figure*}

Current high-altitude airborne camera systems are now able to provide long-term wide-area surveillance videos. To avoid excessive manual effort, it is important that automated systems exist to detect and track objects (mostly vehicles) from the videos. Processing such data is different to traditional object detection problems and has drawn recent attention in both the computer vision and image processing communities. The challenges posed are as follows: lack of appearance information (since objects are always small and usually only grey-level frames are obtained, it is not possible to use appearance information alone to associate objects between frames); well-separated objects (vehicles are almost always on roads, which, in certain parts of the wide-area images, are very distant from one another, making it challenging for any detector that scans an entire image to be computational efficient while not generating large numbers of false alarms); pixel noise (the camera systems are often composed of multiple individual sensors, giving rise to pronounced artifacts, for example related to brightness discontinuities); image registration errors (small errors in image registration can lead to a large number of false alarms).

Mainstream object detection methods broadly sub-divide based on what information is used: spatial information, temporal information or both. For reasons described above, only using spatial information (e.g., the appearance of objects) has proved to be difficult for WAMI detection. There are a number of approaches that consider temporal information\cite{saleemi2013multiframe,LaLondeZS18,SommerTSB16,xiao2010vehicle,prokaj2012tracking}. The simplest approach to using temporal information  would be to perform frame differencing between two consecutive frames. This is straightforward and efficient but will generate two detections for each object. One way to improve this has been to look for the minimum differences among three consecutive frames~\cite{xiao2010vehicle}. Background subtraction goes beyond this idea to build an explicit background model. The current input image is then subtracted from the background to identify foreground objects. \cite{sobral2014comprehensive} reviews background subtraction methods comprehensively. In WAMI videos, the resolution is high and the frame rate low: this can give rise to artifacts caused by parallax, for example. Furthermore, background subtraction methods tend to use lightweight and fast-converging approaches to model the background. For example, \cite{pollard2012detecting} and~\cite{kent2012robust} use modified Gaussian Mixture Models (GMMs) to model each pixel comprising the background while \cite{liang2013vehicle} and \cite{reilly2010detection} compute the mean or median of multiple frames. To reduce the impact of parallax, previous research has considered blocks of pixels in a neighbourhood rather than individual pixels~\cite{xiao2010vehicle} or ignoring regions with large magnitude gradients\cite{liang2013vehicle}.

The benefits of considering both spatial and temporal information has drawn attention in neighbouring contexts: for example, the approach is used in action recognition (e.g.,~\cite{simonyan2014two, wang2016temporal}). There is relatively little research into considering both in the context of WAMI videos. An exception is \cite{LaLondeZS18}, which uses a deep learning based two-stage CNNs approach to generate point detections. The basic idea was to use a lightweight CNN to predict regions that could include moving objects and a deep regression CNN (`FoveaNet') to locate the centres of moving objects (the idea of using a regression network to predict positions directly/indirectly from a single image or multiple frames has been shown to be effective in other applications~\cite{redmon2017yolo9000, rozantsev2017detecting, liu2016ssd}). Moreover, \cite{LaLondeZS18} also considered other combinations (e.g., background subtraction and a single frame foreground object detector\cite{redmon2017yolo9000}), but the performances of these other combinations was not competitive.

In this paper, we focus on developing a processing chain that can detect small moving objects. We consider the second largest images ($15K \times 10K$ pixels) in the WPAFB 2009 dataset~\cite{wpafbdataset}. In this dataset, individual vehicles often occupy less than $100$ pixels. We note that objects of this size are usually assumed to be false alarms in existing papers. We adopt an approach that involves first proposing candidate positions for moving objects and then predicting the objects' positions. We choose to use background subtraction to propose the candidate positions. A very low threshold is used (such that the probability of detection is high), and a small morphological kernel is applied to remove tiny false alarms. We treat the output from the background subtraction as regional proposals and create detection candidates within these areas. An efficient CNN that considers both spatial and temporal information is then trained to reject the false alarms. For the areas with heavy traffic, a lightweight regression network is trained to predict the centre of multiple moving objects from individual detections. This approach of combining the CNNs' and background subtraction's outputs makes it possible for the shape of moving objects to be obtained: this can be useful information that can be exploited by appearance-based tracking systems (e.g., \cite{prokaj2014persistent,chen2017exploring}). Finally, in addition to comparing the detections with the ground-truth, we applied a Gaussian Mixture Probabilistic Hypothesis Density (GM-PHD) filter to the detections to directly utilise the product of the proposed algorithm.

This paper is organised as follows. Section~\ref{sec:imageregistration} briefly reviews the image registration methods that are used in our implementation. The stages comprising the moving object detection algorithm are then described in section~\ref{sec:objectdetection}. The configuration of the GM-PHD filter tracker is introduced in section~\ref{sec:tracking}. The experiment setup and evaluation results are then reported in section~\ref{sec:experiment}. Finally, section~\ref{sec:conclusions} concludes this paper and describes opportunities for future research.

\section{Image Registration} \label{sec:imageregistration}

\begin{figure*}[ht]
\centering
\includegraphics[width=12cm]{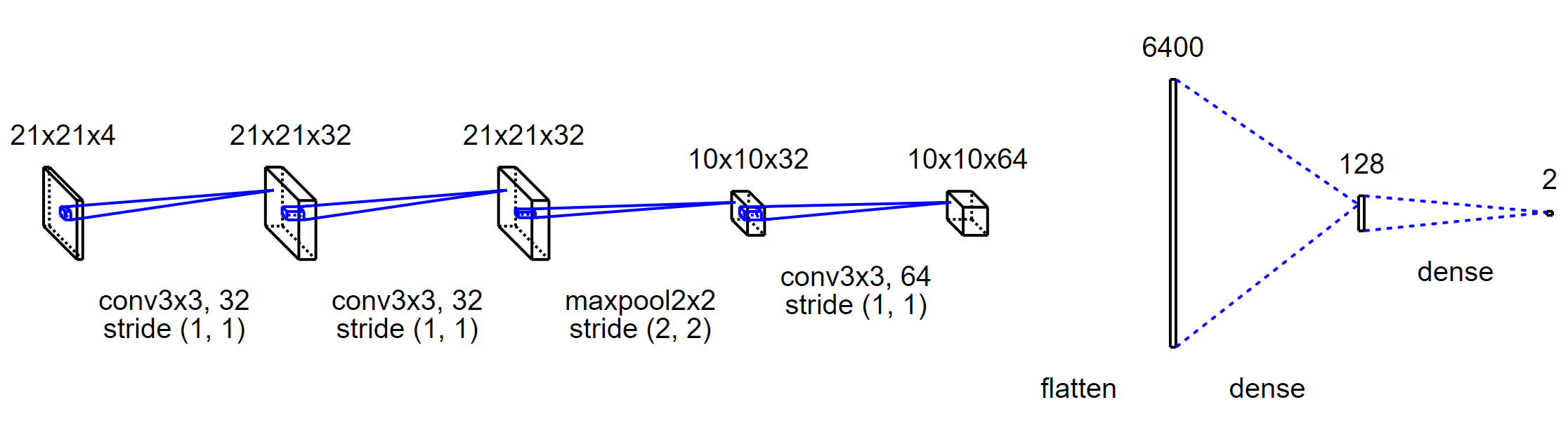}
\caption{The architecture of the classification CNN described in section~\ref{sec:CNN1}. \label{fig:CNN-arch1}}
\end{figure*}

To build a background for the current time from a number of previous frames that were captured by a moving camera system, it is necessary to compensate for the camera motion by aligning all the previous frames to the current frame. This process is also called image registration and the key is to estimate a transformation matrix, $h_{t}^{t-k}$ which denotes transforming from frame $t-k$ to frame $t$, based on a chosen transformation function. In this paper, we use a projective transformation (or Homography) which is widely discussed in multi-perspective geometry, i.e., a closely related area to that in which the WAMI camera system is functioning. 


There are two categories of approach to work out the transformation matrix. Feature-based approaches extract feature points from both images. The feature points can be generated by, for example, Harris corner or SIFT-like~\cite{bay2006surf} detectors. Feature (e.g., SURF, ORB~\cite{rublee2011orb}) descriptors are computed at all the detected feature points. By matching the feature descriptors, pairs of corresponding feature points between two images can be identified and the transformation matrix can then be estimated using RANSAC~\cite{fischler1981random} such that outliers are removed automatically. In our experience, feature-based image registration works well in the context of WAMI video. That said, as it is heavily based on the extracted features, it can sometimes malfunction (e.g., when there are fewer features identified) or the detected features are all concentrated in one area of an image.

The other approach is known as the direct approach and considers the image as a whole. This approach involves directly minimising the difference between the reference and registered images using Lucas-Kanade's algorithm: the implementation is described in~\cite{bouguet2001pyramidal}. Although this approach often has a larger computational cost than feature-based approaches and can struggle if the displacement between two images is large, it typically generates more accurate alignment results and does work more effectively when the number of identified features is insufficient.

\section{Moving Object Detection} \label{sec:objectdetection}

\begin{figure*}[ht]
\centering
\includegraphics[width=12cm]{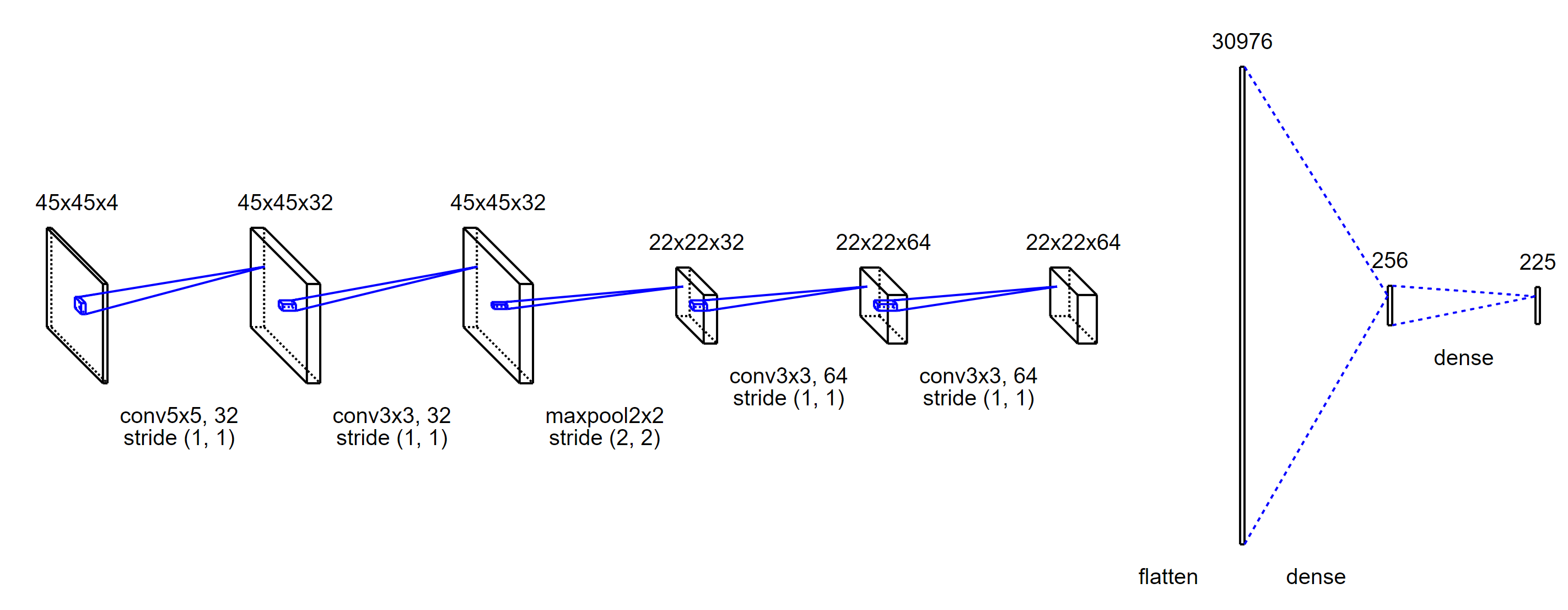}
\caption{The architecture of the regression CNN described in section~\ref{sec:CNN2}.\label{fig:CNN-arch2}}
\end{figure*}

\subsection{Background Subtraction}

We generate the background, $I_{t}^{bg}$, at each time $t$ by computing the median image of the $L$ previous aligned frames. Note that it is unnecessary to perform image registration $L$ times each time a new frame is received. Suppose that we obtained the transformation matrices $h_{t-1}^{t-2}, h_{t-1}^{t-3}, ...$ from processing previous frames, we only need to perform image registration once to get $h_{t}^{t-1}$. Then we can straightforwardly have $h_{t}^{t-2}=h_{t}^{t-1} \cdot h_{t-1}^{t-2}$, $h_{t}^{t-3}=h_{t}^{t-1} \cdot h_{t-1}^{t-3}$ and so on.

The background subtraction result is defined by~$|I_{t}-I_{t}^{bg}| > \tau$ where $\tau$ is the background subtraction threshold. Image morphological operations are applied to remove the tiny blobs that are too small to be an object of interest. As we wish to detect very small (and unclear) objects, the background threshold and minimum blob size are both chosen to be small\footnote{The details of these parameters will be described in table~\ref{tab:bgparams}.}. An example of the output from background subtraction is shown in figure~\ref{fig:example-detector}(b).

In the WPAFB 2009 dataset, it can be observed that there are multiple artifacts at the boundary between different optical sensors (due to the different configurations of the constituent cameras comprising the sensor system). It can also be observed that the brightness of one optical sensor can suddenly change. Both situations influence the result of background subtraction. As suggested in~\cite{KeckGS13}, we use a box filter to compensate for these brightness fluctuations. We perceive that this removes false alarms without reducing the probability of detection significantly.

\subsection{Refining Background Subtraction Output Using a Classification CNN} \label{sec:CNN1}

The major causes of false alarms generated by the background subtraction are: poor image registration, light changes and the apparent displacement of high objects (e.g., buildings and trees) caused by parallax. We emphasise that the objects of interest (e.g., vehicles) mostly appear on roads. More generally, we perceive that a moving object generates a temporal pattern (e.g., road-object-object) that could be exploited to discern whether or not a detection is an object of interest. Thus, in addition to the shape of the vehicle in the current frame, we assert that the historical context of the same place can help to distinguish the objects of interest and false alarms. We therefore create a binary classification CNN to predict if a $21 \times 21$ pixels window contains a moving object given aligned image patches from the previous $N$ frames. We suggest $N=3$ in this paper, because moving vehicles usually take less than four frames (including the current frame) to cross a $21 \times 21$ image patch. The input to the CNN is a $21 \times 21 \times 4$ matrix and the convolutional layers are identical to the traditional 2D CNNs except that the three colour channels are substituted with $N=3$ grey-level frames. The architecture of the proposed CNN is shown in figure~\ref{fig:CNN-arch1}. A batch normalisation layer~\cite{IoffeS15} is added after each convolutional layer and fully connected layer. A softmax layer is applied to output to obtain probability-like scores.

Background subtraction is used as a regional proposal method, however the connected components are sometimes too small or too large with respect to the ground-truth (moving) objects. To extract the regions to be used by the CNN, we subdivide the images into $5 \times 5$ cells. If a cell contains any pixels that are classified as a candidate detection by the background subtraction, we pick the $21 \times 21$ window centred on this cell as the input to the CNN (see figure~\ref{fig:example-detector}(c)). If the CNN classifies the windows as positive, this cell will be flagged as containing a moving object (see figure~\ref{fig:example-detector}(d)).

The proposed CNN is trained as follows. For each frame in the training set, the background subtraction is used to propose cells. All the cells that have their centres within the range of $6$ pixels from any ground-truth points are used as positive samples. All the cells that are at least $15$ pixels away from all the ground-truth points are used as negative samples. We note that the positive samples are dependent on the output of background subtraction, such that the obscured and subtle detections (i.e., potential objects whose appearances are too similar to the background) are not considered in training. We do this because we want to couple the CNN to the regional proposal method. This means that the trained model becomes specific to the configuration of the background subtraction process. It transpires that there are many more negative samples than positive ones. Indeed, there are sufficiently many negative samples that we cannot train the model with all the extracted negative samples. We therefore use a negative set which is four times larger than the positive set to train the CNN. Since the training negative set is then much smaller than the complete negative set, the threshold for accepting detections (as~(\ref{eq:CNN1output})), $\phi$, was adjusted by a validation set to maximise the $F_{1}$ score.
\begin{equation}
prediction=\left\{
\begin{array}{cc}
true & o_{1} >= \phi \\
false & o_{2} >= \phi
\end{array} \right.
\label{eq:CNN1output}
\end{equation}
\noindent where $o_{1}$ and $o_{2}$ are the outputs of the binary CNN and $o_{1}+o_{2}=1$ due to the softmax layer. In traditional CNNs, $\phi=0.5$.

\subsection{Predicting Object Positions Using a Regression CNN} \label{sec:CNN2}

As illustrated in figure~\ref{fig:example-detector}(b), two major problems that are present in the background subtraction output include the existence of multiple detections for one object and of one merged detection for multiple objects. While the former problem can be addressed by considering morphological operations, this makes the latter problem worse. Thus, it is necessary to separate the detections into two sets: detections which contain a single object that can be outputted directly and detections that may contain multiple objects. First, we assign the candidate detections output by the background subtraction to the accepted detection regions (figure~\ref{fig:example-detector}(d)) and remove the candidate detections without any assignment. We refer the accepted blobs from background subtraction as $S_{bg}=\{s^{bg}_{1}, s^{bg}_{2}, ...\}$, and the blobs from classification CNN as $S_{cnn}=\{s^{cnn}_{1}, s^{cnn}_{2}, ...\}$. The candidate detection $s^{cnn}_{i}$ which can be directly outputted must satisfy the following two conditions, otherwise we consider there might be multiple objects giving rise to the single candidate detection:
\begin{itemize}
\item $s^{bg}_{j}$ is assigned to $s^{cnn}_{i}$ and no other $s^{bg}_{k}$ with $j \ne k$ is assigned to $s^{cnn}_{i}$.
\item The size of $s^{cnn}_{i}$ is smaller than $150$ pixels.
\end{itemize}

The candidate detections can be accepted when the common area between the assigned pair, $s^{bg}_{i}$ and $s^{cnn}_{j}$, is larger than $50$. This is such that we take full advantage of the background subtraction algorithm. An illustration of the two sets is shown in figure~\ref{fig:example-detector}(e).

\begin{figure}[h]
\centering
\subfigure[The input cubic to the regression CNN. From left to right: $t$, $t-1$, $t-2$, $t-3$]{
\includegraphics[width=5cm]{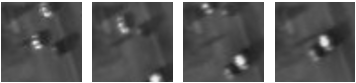}
}
\par
\subfigure[The response (resized to $45 \times 45$) of the regression CNN given the above input.]{
\includegraphics[width=3cm]{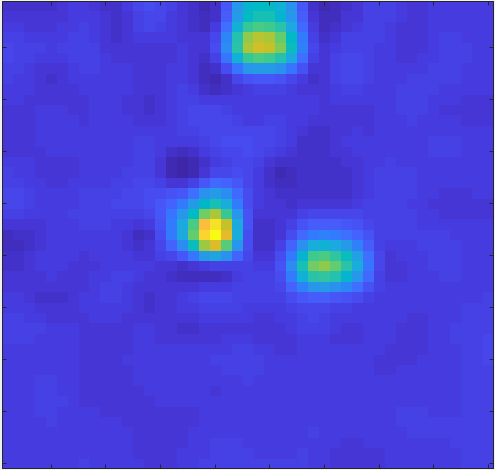}
}
\caption{An example of how the regression CNN works. \label{fig:example-regression}}
\end{figure}

As proposed in~\cite{LaLondeZS18}, a regression CNN can predict the positions of objects given spatial and temporal information. Such a second CNN can be used to deal with the set of detections that may contain multiple detections. We therefore train a regression CNN, whose architecture is shown in figure~\ref{fig:CNN-arch2}, to predict the positions of moving objects. The input to this CNN is similar to the classification CNN described in section~\ref{sec:CNN1} but the size of the patches changes to $45 \times 45$ (as shown in figure~\ref{fig:example-regression}(a)). The response of the CNN is a $225$ dimensional vector, equivalent to a down-sampled image ($15 \times 15$) for reducing computational cost. For simplification, we always up-sample the response from the CNN to a $45 \times 45$ image in the subsequent discussions. An example of the CNN response is shown as figure~\ref{fig:example-regression}(b).

While preparing the training set, for each frame, we obtain a number of $45 \times 45 \times 4$ data-cubes. We assume that $\{(x_{1}^{g}, y_{1}^{g}), (x_{2}^{g}, y_{2}^{g}), ..., (x_{M}^{g}, y_{M}^{g}) \}$ denotes the set of targets in the ground-truth for current frame, where $M$ is the number of targets. If we only use these ground-truth points as centres of windows to extract the data-cubes, a moving object could always be located in the centre of every window (which would bias the regression model). So, some drifts are added to each of the ground-truth points: $C_{m}=(x_{m}^{g}+\Delta x_{m}, y_{m}^{g}+\Delta y_{m})$ where $\Delta x_{m}$ and $\Delta y_{m}$ are each drawn from a uniform distribution on $[-8,8]$. In order to get a more generalised training set, we iterate this process five times creating $C_{m,1}, ... C_{m,5}$. Such that, for each ground-truth detection, we create five data-cubes and each data-cube contains the patches of the current frame and (aligned) three previous frames centred on $C_{m, l}$ where $l={1, ..., 5}$. The training set of the CNN response is thus straightforward. We down-sample the $45 \times 45$ current frame patch to $15 \times 15$. If there is a ground-truth target within a pixel, the corresponding pixel in the CNN response is set to $1$ otherwise $0$. Therefore, the responses for training are either $0$ or $1$.

Although the training response is $0$ or $1$, the actual response from the regression is between $0$ and $1$; see, for example, figure~\ref{fig:example-regression}(b). We cannot extract the positions by directly picking out the positions wherever the CNN response value is $1$, because a weak detection can often give outputs that are less than $1$. So, after setting a minimum value ($\kappa$) for the possible pixels, we consider the peaks in the response image as the centres of detections. A default bounding box will be given if the detection is made by the regression CNN. The detection results of both CNNs are illustrated in figure~\ref{fig:example-detector}(f).

\section{Tracking} \label{sec:tracking}

In this section, we focus on applying a multi-target tracker to generate multiple tracks using the output from the proposed detector. In order to reflect the performance of the detector directly, we only use the centres of detections (i.e., we do not use objects' appearance at this stage). The Gaussian Mixture Probabilistic Hypothesis (GM-PHD) filter~\cite{vo2006gaussian} is used to process the detections and an implementation can be found in~\cite{gmphdimple}. We consider the near-constant velocity model as the dynamic model for the objects. The transition function is as standard and process noise is defined as (\ref{eq:processnoise}). Since the position of the detections are always with respect to the current frame, the states are compensated for the camera motion as well.
\begin{equation}
\epsilon_{t} \thicksim \mathcal{N}(0, Q)
\label{eq:processnoise}
\end{equation}
\noindent and $Q$ is defined as follow:
\begin{equation}
Q = \sigma_{q}^{2} \cdot \begin{bmatrix}
   \frac{1}{3} dt^{3} & 0 & \frac{1}{2} dt^{2} & 0 \\
   0 & \frac{1}{3} dt^{3} & 0 & \frac{1}{2} dt^{2} \\
   \frac{1}{2} dt^{2} & 0 & 1 & 0 \\
   0 &\frac{1}{2} dt^{2} & 0 & 1 \\
   \end{bmatrix}
\end{equation}
\noindent where $\sigma_{q}$ is the magnitude of the process noise which should be adjusted based on the video content and $dt$ is the time interval.

A track is initialised if a pair of points, $p_{t|t-1}=[p_{t|t-1}^x, p_{t|t-1}^y]$ and $p_{t}=[p_{t}^x, p_{t}^y]$ can be found such that $dist(p_{t|t-1}, p_{t}) < \theta$, where $p_{t}$ is a detection position at frame $t$, $p_{t|t-1}$ is a detection position which is observed in frame $t-1$ and transformed to frame $t$. The state of the initialised track is defined as $s_{t}=\left[p_{t}^x, p_{t}^y, p_{t}^x-p_{t|t-1}^x, p_{t}^y-p_{t|t-1}^y\right]^{T}$ and the initial weight for each Gaussian is $\omega_{init}$. We note that there can be a number of tracks initialised that duplicate existing tracks. These are typically merged as part of the pruning process within the GM-PHD. A track is confirmed if its weight is larger than $\omega_{show}$. A track will be removed if its weight is lower than $\omega_{remove}$ or moves out of the image.

A list of recommended parameters for the GM-PHD filter regarding WPAFB 2009 dataset is shown in table~\ref{tab:phdparams}.

\begin{table}[h]
\centering
\begin{tabular}{c|c}
\hline
Parameter description & Value \\
\hline
Time interval ($dt$) & 1 \\
Process noise ($\sigma$) & 3 \\
Measurement noise ($R$) & 3 \\
Maximum velocity ($\theta$) & 35 \\
Birth weight ($\omega_{init}$) & 0.25 \\
Extraction weight ($\omega_{show}$) & 0.5 \\
Remove weight ($\omega_{remove}$) &  0.05 \\
Probability of detection & 0.8 \\
Probability of Survival & 0.95 \\
\hline
\end{tabular}
\caption{The recommended parameters for the GM-PHD filter for WPAFB 2009 dataset. \label{tab:phdparams}}
\vspace{-5mm}
\end{table}

\section{Experimental Results} \label{sec:experiment}

We used the WPAFB 2009~\cite{wpafbdataset} dataset to train and evaluate the proposed approach. The images were taken by a camera system with six optical sensors and had already been stitched to cover a wide area of around $35\, km^{2}$. This dataset includes 1025 frames and is divided into training video ($512$ frames) and test video ($513$ frames). All the vehicles and their trajectories are manually annotated. There are multiple resolutions of videos in the dataset. Unlike most of previous papers, which consider the largest ones ($25K\times20K$ pixels), we chose to use the smaller ones ($12K\times10K$ pixels) where the size of vehicles are often smaller than $10 \times 10$ pixels.

Because the proposed algorithm is focusing on detecting moving objects, we remove all the objects with insufficient movement from the ground-truth for evaluation (this was also the approach to evaluation considered in \cite{SommerTSB16, LaLondeZS18}). An object whose displacement is smaller than $0.8~metres$ (as calculated from the latitudes and longitudes provided in the ground-truth) between two consecutive frames is removed. This scenario reduces the size of ground-truth from 1,176,447 to 455,427 true detections in the training video and from 1,145,779 to 460,386 in the test video.

We used the first 300 frames in the training video as the training set for both CNNs. We randomly picked 500,000 positive samples and 2,000,000 negative samples to train the classification CNN. We then validated this CNN with the same 300 frames to obtain $\phi$ (as described in section~\ref{sec:CNN1}). The regression CNN was trained using 500,000 samples that were extracted via the process described in section~\ref{sec:CNN2}.

\begin{figure}[h]
\centering
\includegraphics[width=7cm]{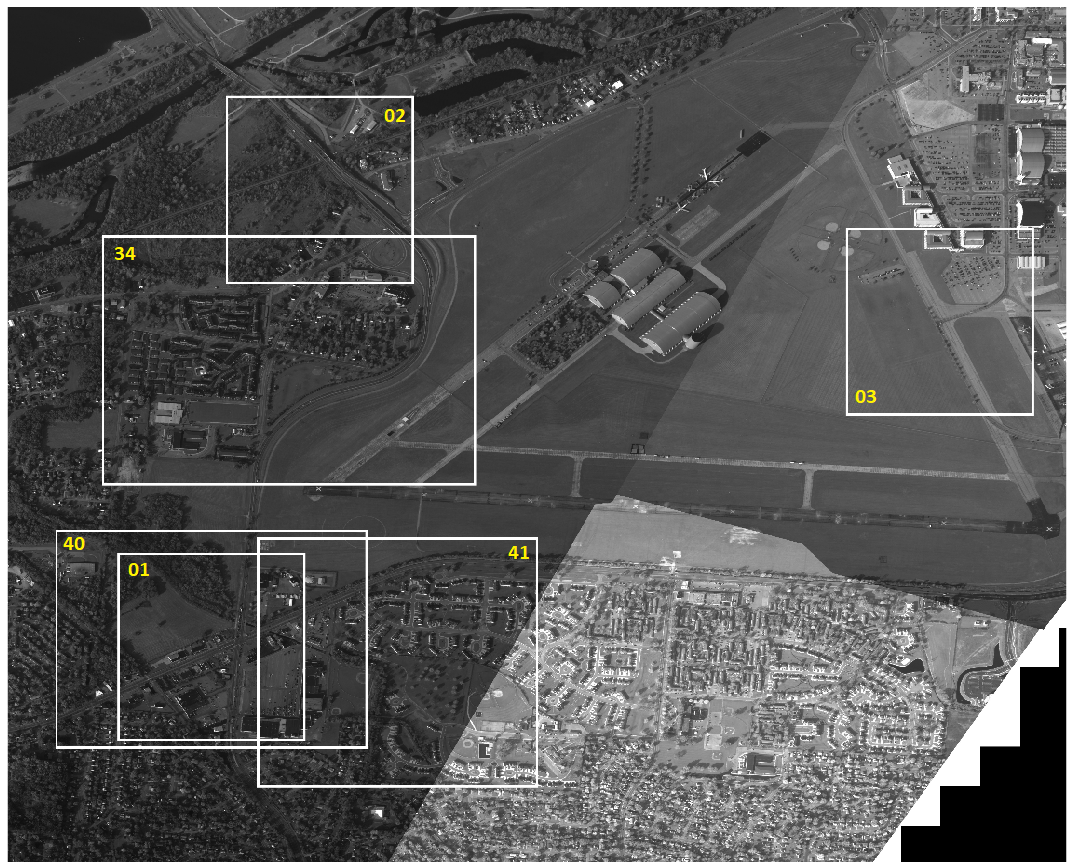}
\caption{The areas of interest that are used in our experiments. \label{fig:imagerois}}
\end{figure}

We tested the proposed detector on the full stitched images in both the training and test videos. Feature-based image registration (see section~\ref{sec:imageregistration}) was used to align consecutive frames. To compare both detection and tracking performance with existing papers, we additionally create six sub-videos including different Areas Of Interest (AOI 01, 02, 03, 34, 40, 41 in \cite{SommerTSB16, LaLondeZS18, BasharatTXASFTH14, KeckGS13}). As all these papers did not fully specify the details about the areas of interest, we use our pre-processing architecture to generate sub-videos which attempt to cover similar regions to those specified in these papers. The areas of interest in our experiment are shown in figure~\ref{fig:imagerois}. By using the image registration results for the full images, we can produce a video that is centred on a particular point. Thus there will be no vertical or horizontal translations over time, even though rotations and scalings are apparent in the sub-videos. To process the sub-videos, we use the direct approach (see section~\ref{sec:imageregistration}) to perform image registration, since processing time is acceptable regarding the lower resolution of AOIs\footnote{We noticed that for some sub-videos, the feature-based image registration was not working well as the extracted features were not well distributed across the image. In the other sub-videos, the influence of the choice of registration on detector performance was negligible.}.

We used different parameters in background subtraction on the full stitched images and the AOIs since pronounced noise was present in some areas. The parameters are shown in table~\ref{tab:bgparams}.

 \begin{table}[h]
\centering
\begin{tabular}{m{5cm}|m{1cm}|m{1cm}}
\hline
Parameter & Full & AOI \\
\hline
Number of previous frames ($L$) & 3 & 5 \\
\hline
Background subtraction threshold ($\tau$) & 8 & 5 \\
\hline
Image opening kernel & $3 \times 3$ & $3 \times 3$ \\
\hline
Threshold for accepting detections in classification CNN ($\phi$) & 0.8 & 0.8 \\
\hline
Min. response in Regression CNN ($\kappa$) & 0.25 & 0.25 \\
\hline
\end{tabular}
\caption{The parameters used while processing the full stitched images and the areas of interest (AOI). \label{tab:bgparams}}
\vspace{-5mm}
\end{table}

\subsection{Evaluation Method}

\begin{table*}[h]
\centering
\begin{tabular}{c|c|c|c|c|c|c|c}
\hline
Metric & 01 & 02 & 03 & 34 & 40 & 41 & full image \\
\hline
 & \multicolumn{7}{c}{Training video} \\
\hline
Precision & 0.953 & 0.965 & 0.937 & 0.970 & 0.955 & 0.960 & 0.952 \\
Recall & 0.932 & 0.951 & 0.964 & 0.961 & 0.931 & 0.928 & 0.899 \\
\hline
 & \multicolumn{7}{c}{Test video} \\
 \hline
Precision & 0.939 & 0.956 & 0.934 & 0.957 & 0.942 & 0.943 & 0.941 \\
Recall & 0.931 & 0.936 & 0.958 & 0.949 & 0.928 & 0.925 & 0.891 \\
\hline
\end{tabular}
\caption{The precisions and recalls on different areas of interest using the proposed detector.} \label{tab:result-1}
\vspace{-5mm}
\end{table*}

The performances of the detector and the tracker were considered separately. For \textit{measuring the detector}, we define the metrics as follows (in the same way as was considered elsewhere\cite{LaLondeZS18, SommerTSB16}): A detection is considered as true positive when there is at least one ground-truth point within $10$~pixels range and each ground-truth point can only be assigned to one detection. All the other detections are false positives. The false negative set includes the ground-truth points for which there is no detection within the range of $10$~pixels. All the unassigned detections and ground-truth points were marked as false alarms and mis-detections respectively. The precision, recall and $F_1$ score are using the standard equations, which due to space limitations are not explicitly defined here.





In order to \textit{measure the tracker}, we used the definitions and metrics in \cite{KeckGS13, BasharatTXASFTH14}. The \textbf{estimated track will be referred to as `track'} and the \textbf{ground-truth target trajectory will be referred as `target trajectory'} in the subsequent discussions. For each target trajectory, we remove the way-points that occur when the objects is almost stationary, but we do not separate it into tracklets. We also ignore all the very short estimated tracks and targets trajectories (less than $5$ frames) as they may not be initialised and tracked well in practice. To calculate \textbf{Target purity}, we assign the most similar tracks to each target trajectory and compute the percentage of the pre-dominate track. \textbf{Target continuity} is defined as the number of tracks that are assigned to one target trajectory\footnote{For example, in a case that the tracker generates three tracks which are 5, 26 and 64 frames long, and they can be assigned to a target trajectory which moves for 40 frames, stops for 25 frames and then moves for 60 frames, the purity is $64/100=64\%$ and the continuity is $2$.}. We report the average values over all the target trajectories. These metrics evaluate the fragmentation of the tracks. \textbf{Track purity} and \textbf{Track continuity} are otherwise centred on the tracks: for each track we assign target trajectories to it and calculate the metrics in the same way as mentioned above. `Track purity \& continuity' present how much one track includes multiple targets. Again, the average over all the tracks is reported. The range of purity is $[0, 100\%]$ with $100\%$ being the ideal. The continuity is within $[1, \infty]$, with $1$ being ideal.

\subsection{Detection Results}

The precision and recall of the proposed detector when applied to different AOIs and the full image are presented in table~\ref{tab:result-1}. We can see that the performance on the training video is (as expected) better than on the test video. We believe this is due to the relatively small number of negative samples and because the threshold for accepting detections, $\phi$, is estimated using on training set. Most of the AOIs yield similar precisions, except for AOI 03 which is lower because there are some non-vehicle moving objects that are detected by the algorithm, but were not included in the ground-truth. In terms of missed detections, the main cause we have identified is that some moving objects cannot be detected by background subtraction: for example, when they look similar to the background and move slowly. Regarding the results on the full images, the lower recall is caused by a larger background subtraction threshold $c$, which leads to a smaller number of more reliable detections.

Table~\ref{tab:detectioncompare} presents the $F_{1}$ score of the proposed algorithm compared to the best detectors we are aware of in the literature. Given that we are considering smaller objects, it is evident that the proposed detector is competitive with the others.

\begin{table}[h]
\centering
\begin{tabular}{c|c|c|c|c|c|c}
\hline
Methods & 01 & 02 & 03 & 34 & 40 & 41 \\
\hline
\cite{LaLondeZS18} & 0.947 & 0.951 & 0.942 & 0.933 & 0.983 & 0.928 \\
\cite{SommerTSB16} & 0.866 & 0.890 & 0.900 & - & - & - \\
\cite{TeutschG16} & - & - & - & 0.874 & 0.847 & 0.854 \\
\hline
Ours(train) & 0.942 & 0.958 & 0.950 & 0.965 & 0.943 & 0.944 \\
Ours(test) & 0.935 & 0.947 & 0.945 & 0.953 & 0.935 & 0.934 \\
\hline
\end{tabular}
\caption{The $F_1$ scores of existing algorithms (as assessed on larger targets only) and of our approach on the AOIs.\label{tab:detectioncompare}}
\vspace{-5mm}
\end{table}

\subsection{Tracking Results}

The performance of applying a GM-PHD filter to the detections is reported in table~\ref{tab:trackingresults}. We observe that AOI 01 and 40 are most challenging because in the middle of the area there is a crowded crossroads where vehicles start and stop frequently. AOI 34 includes heavy traffic and vehicles are very close to each other. The traffic is light in AOI 03, but there are a couple of traffic lights which cause the tracks to fragment. The tracker performs similarly in AOI 34 and AOI 03. The tracker achieves good results in AOI 02 and 41 where traffic is moderate and no traffic lights are present.

In comparison with \cite{KeckGS13} and~\cite{BasharatTXASFTH14}, the advantage of our approach is obvious in heavy traffic. In AOI 34 and 41, the target purities from \cite{BasharatTXASFTH14} were below $50 \%$ and even reduced to $28 \%$ in AOI 40. In AOI 02, our approach yields much higher target purity compared to~\cite{KeckGS13} which is below $20 \%$. However, in areas with light traffic such as AOI 03, our approach yields similar results to~\cite{KeckGS13}. In terms of track purity and continuity, the proposed approach outperforms the others as well: our estimated tracks each contain fewer targets on average.

\begin{table}[h]
\centering
\begin{tabular}{c|c|c|c|c|c|c}
\hline
Metric & 01 & 02 & 03 & 34 & 40 & 41 \\
\hline
Target Puri. (\%) 	& 53.1 	& 71.7 	& 64.7 		& 64.9 	& 52.2 		& 70.9 \\
Target Cont. 	& 2.23 		& 1.62 		& 1.92 			& 1.98	 	& 2.46		 	& 1.46 \\
Track Puri. (\%) 	& 89.4 	& 93.4 	& 90.64 	& 92.0 	& 89.5	 	& 93.7 \\
Track Cont. 	& 1.10 		& 1.07 		& 1.07 			& 1.06	 	& 1.11		 & 1.03 \\
\hline
\end{tabular}
\caption{The performance of GM-PHD filter using the proposed detector on AOIs of the WPAFB2009 (test video).\label{tab:trackingresults}}
\vspace{-5mm}
\end{table}

\section{Conclusions} \label{sec:conclusions}

In this paper, we propose a moving object detector for wide-area motion imagery (WAMI) videos. The detector is based on using background subtraction with a low threshold to ensure large numbers of potential detections are identified. False alarms are removed and merged detections disentangled using two CNNs that both consider spatial-temporal information. Our experimental results shows competitive performance compared to the state-of-the-art while dealing with smaller objects. We also demonstrate the feasibility of applying multi-target track to the detections we generate.

Future research includes: assessing the extent to which the trained CNNs can be applied to other WAMI videos; developing techniques to detect stationary vehicles in areas identified from an inability to detect a tracked object; tracking that can persist across long baselines by exploiting the appearance information derived from the background subtraction process.

\section*{Acknowlegment}

This work was funded through ``Track Analytics For Effective Triage Of Wide Area Surveillance Data'' by the Defence Science \& Technology Laboratory (DSTL).

\bibliographystyle{ieeetr}
\bibliography{reference}

\end{document}